\pdfoutput=1

\documentclass[11pt]{article}
\PassOptionsToPackage{table,dvipsnames}{xcolor}

\usepackage[preprint]{acl}

\usepackage{amssymb}
\usepackage{pifont}
\usepackage{graphicx}
\usepackage{booktabs}
\usepackage{fp}
\usepackage{colortbl}
\usepackage{pgfplots}
\usepackage{tikz}
\usepackage{colortbl}
\usepackage{graphicx}
\usepackage{booktabs}
\usepackage{makecell}
\usepackage{listings}
\usepackage{xcolor} 

\lstdefinestyle{pythonstyle}{
    language=Python,
    basicstyle=\ttfamily\small,
    keywordstyle=\color{blue},
    stringstyle=\color{green!50!black},
    commentstyle=\color{gray},
    showstringspaces=false,
    breaklines=true,
    frame=single,
    columns=fullflexible
}
\newcommand{\cmark}{\ding{51}} 
\newcommand{\xmark}{\ding{55}} 

\usepackage[table,dvipsnames]{xcolor} 
\usepackage{pgf}
\usepackage{collcell}
\usepackage{booktabs}
\usepackage{quoting}

\newcommand*{\MinNumber}{0.02}%
\newcommand*{\MaxNumber}{0.4}%

\colorlet{lgreen}{BrickRed!9}

\newcommand{\heatcell}[1]{%
  \pgfmathparse{int(100*(#1-\MinNumber)/(\MaxNumber-\MinNumber))}%
  \edef\hmvalue{\pgfmathresult}
  \edef\heatmapcellcolor{BrickRed!\hmvalue!lgreen}
  \expandafter\cellcolor\expandafter{\heatmapcellcolor}%
  \pgfmathparse{#1 > 0.5 ? 1 : 0}%
  \ifnum\pgfmathresult=1
    \textcolor{white}{#1}%
  \else
    \textcolor{black}{#1}%
  \fi%
}

\usepackage{times}
\usepackage{latexsym}
\usepackage{amssymb}
\usepackage{bbm}
\usepackage{amsmath}
\usepackage{dsfont}
\usepackage{booktabs}
\usepackage{multirow}
\usepackage{colortbl}
\usepackage{colortbl}  
\usepackage{booktabs}   

\expandafter\def\expandafter\normalsize\expandafter{%
    \normalsize%
    \setlength\abovedisplayskip{1pt}%
    \setlength\belowdisplayskip{1pt}%
}

\usepackage[T1]{fontenc}
\usepackage[utf8]{inputenc}
\usepackage{microtype}
\usepackage{inconsolata}
\usepackage{graphicx}

\usepackage{geometry}
\usepackage{enumitem}
\usepackage{fancybox}

\title{TruthTorchLM:\\ A Comprehensive Library for Predicting Truthfulness in LLM Outputs}

\author{
    Duygu Nur Yaldiz\textsuperscript{1}\footnotemark[1] \quad 
   Yavuz Bakman\textsuperscript{1}\footnotemark[1] \quad Sungmin Kang\textsuperscript{1}   \\
   \bf \quad Alperen Ozis\textsuperscript{2} \quad
   Hayrettin Eren Yildiz\textsuperscript{3} \quad Mitash Shah \textsuperscript{1} \\
   \bf Zhiqi Huang \textsuperscript{4} \quad Anoop Kumar\textsuperscript{4} \quad
   Alfy Samuel\textsuperscript{4} \quad Daben Liu\textsuperscript{4}  \\
   \bf Sai Praneeth Karimireddy\textsuperscript{1} \quad
   Salman Avestimehr\textsuperscript{1}\\
\textsuperscript{1}University of Southern California \quad
  \textsuperscript{2}Independent Researcher \\ \textsuperscript{3}Bogazici University \quad \textsuperscript{4} Capital One \\
  \texttt{\{yaldiz, ybakman\}@usc.edu}}

\begin{document}

\maketitle
\begin{abstract}

Generative Large Language Models (LLMs) inevitably produce untruthful responses. Accurately predicting the truthfulness of these outputs is critical, especially in high-stakes settings. To accelerate research in this domain and make truthfulness prediction methods more accessible, we introduce \texttt{TruthTorchLM}\footnote{\url{https://github.com/Ybakman/TruthTorchLM}},\footnote{\url{https://www.youtube.com/watch?v=dgovBgUYz3w}},\footnote{\url{https://pypi.org/project/TruthTorchLM/}} an open-source, comprehensive Python library featuring over 30 truthfulness prediction methods, which we refer to as \textit{Truth Methods}. Unlike existing toolkits such as \texttt{Guardrails} \cite{guardrails2024}, which focus solely on document-grounded verification, or \texttt{LM-Polygraph} \cite{fadeeva-etal-2023-lm}, which is limited to uncertainty-based methods, \texttt{TruthTorchLM} offers a broad and extensible collection of techniques. These methods span diverse trade-offs in computational cost, access level (e.g., black-box vs. white-box), grounding document requirements, and supervision type (self-supervised or supervised). \texttt{TruthTorchLM} is seamlessly compatible with both HuggingFace and LiteLLM, enabling support for locally hosted and API-based models. It also provides a unified interface for generation, evaluation, calibration, and long-form truthfulness prediction, along with a flexible framework for extending the library with new methods. We conduct an evaluation of representative truth methods on three datasets, TriviaQA, GSM8K, and FactScore-Bio.

\end{abstract}

\section{Introduction}
Generative Large Language Models (LLMs) have been widely adopted in many real-world applications due to their remarkable performance across a range of tasks, from code generation to conversational agents~\cite{band2022benchmarking}. Despite these successes, LLMs inevitably produce outputs that are factually or logically incorrect, commonly referred to as \emph{hallucinations}~\cite{ravi2024lynxopensourcehallucination}. Detecting such untruthful outputs is particularly crucial in high-stakes applications where reliability and correctness are essential.

In response, numerous methods have been proposed to assess the truthfulness of LLM-generated content to support reliable decision-making. These include uncertainty estimation techniques, agentic tool use, multi-LLM collaboration strategies, supervised classification models, and document-based verification approaches. Each method varies in terms of computational cost, required access to model internals, and reliance on external resources. As LLM usage continues to grow, developing new techniques and refining existing ones remains crucial, given that truthfulness is a core requirement for trustworthy language generation.

\begin{figure*}
\vskip -0.2in
\begin{center}
\includegraphics[width=0.97\textwidth]{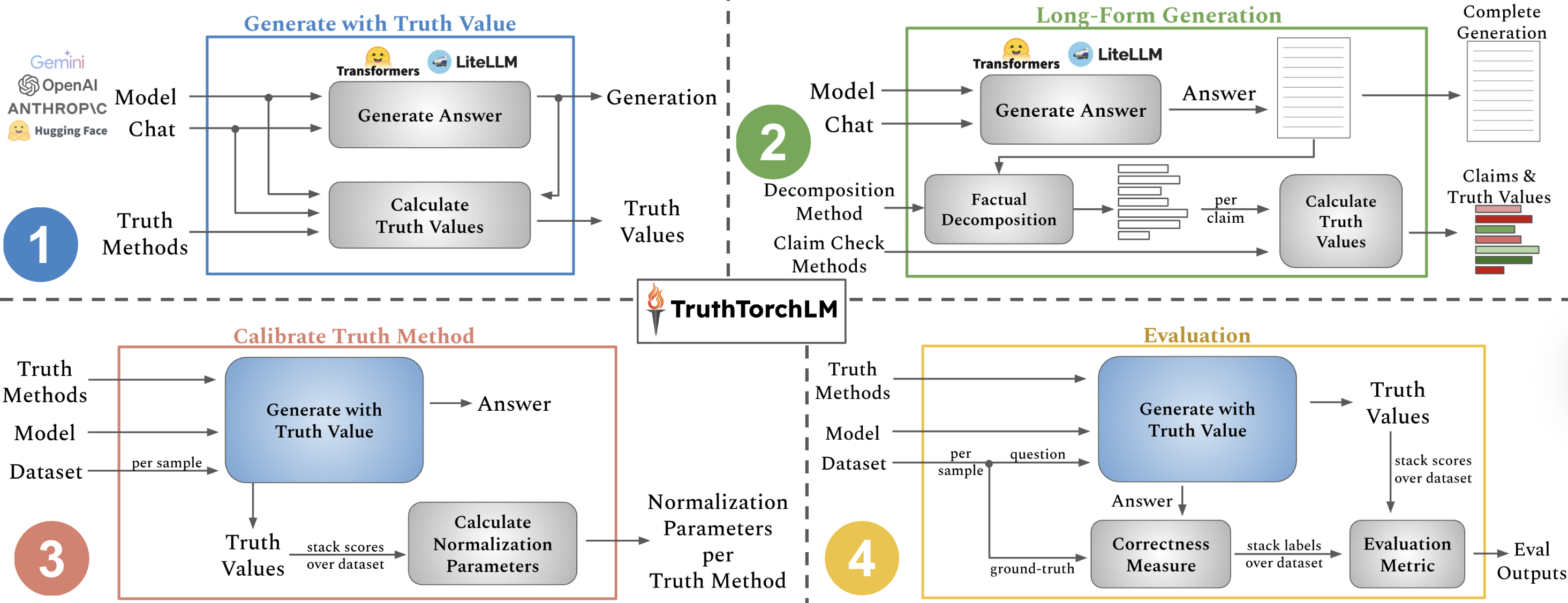}
\vskip -0.1in
\caption{Overview of TruthTorchLM functionalities.}
\label{fig:overview}
\end{center}
\vskip -0.25in
\end{figure*}

To support research in this domain, an open-source library that consolidates existing methods and offers a flexible development framework is essential. Current software tools only partially meet this need. For instance, \texttt{Guardrails} \cite{guardrails2024} is an open-source library that focuses on document-based guardrails to assess the truthfulness of LLM outputs. However, it lacks support for a broader range of methods that do not rely on external documents, such as uncertainty estimation techniques. Similarly, \texttt{LM-Polygraph} \cite{fadeeva-etal-2023-lm} provides implementations of uncertainty estimation methods, but it does not include supervised approaches, document-checking techniques, or tool-based strategies. A more comprehensive and extensible toolkit is needed to facilitate systematic evaluation and innovation across the full spectrum of truthfulness prediction methods.

To facilitate research and address the limitations of existing software in the domain of truthfulness prediction, we introduce \texttt{TruthTorchLM }(TTLM), an open-source library that currently implements over 30 \textit{truth methods} with diverse algorithmic ideas. The library provides an intuitive and extensible interface for integrating new methods, enabling researchers to prototype and evaluate novel approaches with ease. \texttt{TruthTorchLM} is seamlessly compatible with both HuggingFace~\cite{wolf-etal-2020-transformers} and LiteLLM~\cite{litellm2024}, two of the most widely used frameworks for deploying LLMs in local and hosted environments. Beyond method implementation, the library offers comprehensive evaluation tools for benchmarking performance and includes calibration utilities to produce more interpretable truthfulness scores. Importantly, \texttt{TruthTorchLM} supports the application of truth methods to long-form generations, where multiple factual claims may be present and each claim requires individual assessment. This long-form setting represents a challenging and underexplored area of research, which is currently underserved by existing libraries.

Our contributions can be summarized as follows:
\begin{enumerate}[topsep=0.5pt]
\itemsep -0.08in
    \item We release \texttt{TruthTorchLM (TTLM)}, an open-source library that implements over 30 truthfulness prediction methods, fully compatible with both HuggingFace and LiteLLM frameworks.
    
    \item \texttt{TruthTorchLM} provides a unified interface for generation, evaluation, calibration, and long-form extension of existing truth methods, along with a flexible framework for adding new methods.
    
    \item We conduct a comprehensive evaluation of a representative truth methods across three diverse datasets, TriviaQA, GSM8K, and FactScore-Bio, using both an open-weight model (LLaMA-3-8B) and a closed-weight model (GPT-4o-mini).
\end{enumerate}


    

\section{System Design and Features of TTLM}
\texttt{TTLM}  is designed around a central abstraction: \textit{truth methods}, which are methods for predicting the truthfulness of LLM-generated outputs. Using TTLM, users can generate responses for any input query and apply one or more truth methods to assess the reliability of these outputs, whether they are short-form answers containing a single claim or long-form responses with multiple factual assertions. In addition to prediction, TTLM enables users to evaluate and calibrate the outputs of truth methods with just a few lines of code. In the following sections, we detail each of TTLM’s core features and explain how they support robust and scalable research in truthfulness assessment.

\begin{table*}[h!]
\centering
\caption{Categorization of a representative subset of available methods in TruthTorchLM.}
\vskip -0.15in
\small 
\setlength{\tabcolsep}{4pt} 
\begin{tabular}{lcccc}
\toprule
\textbf{Truth Methods} & \textbf{Document-Grounding} & \textbf{Supervised} & \textbf{Access Level} & \textbf{Sampling-Required} \\
\midrule
LARS \cite{yaldiz2024designlearntrainablescoring} & \xmark & \cmark & Grey-box & \xmark \\
MARS \cite{bakman2024mars} & \xmark & \xmark & Grey-box & \xmark \\
SelfDetection \cite{zhao-etal-2024-knowing} & \xmark & \xmark & Black-box & \cmark \\
PTrue\cite{kadavath2022language} & \xmark & \xmark & Grey-box & \xmark \\
AttentionScore \cite{sriramanan2024llmcheck} & \xmark & \xmark & White-box & \xmark \\
CrossExamination \cite{cohen-etal-2023-lm} & \xmark & \xmark & Black-box & \cmark \\

Eccentricity \cite{lin2023generating} & \xmark & \xmark & Black-box & \cmark \\
GoogleSearchCheck \cite{chern2023factool} & \cmark & \xmark & Black-box & \xmark \\
Inside \cite{chen2024inside} & \xmark & \xmark & White-box & \cmark \\
KernelLanguageEntropy \cite{nikitin2024kernel} & \xmark & \xmark & Black-box & \cmark \\
MiniCheck \cite{tang-etal-2024-minicheck} & \cmark & \xmark & Black-box & \xmark \\
Matrix-Degree \cite{lin2023generating} & \xmark & \xmark & Black-box & \cmark \\
SAPLMA \cite{azaria-mitchell-2023-internal} & \xmark & \cmark & White-box & \xmark \\
SemanticEntropy \cite{kuhn2023semantic} & \xmark & \xmark & Grey-box & \cmark \\
MultiLLMCollab \cite{feng-etal-2024-dont} & \xmark & \xmark & Black-box & \cmark \\
SAR \cite{tokensar} & \xmark & \xmark & Grey-box & \cmark \\
VerbalizedConfidence \cite{tian-etal-2023-just} & \xmark & \xmark & Black-box & \xmark \\
DirectionalEntailmentGraph \cite{da2024llmuncertaintyquantificationdirectional} & \xmark & \xmark & Black-box & \cmark \\
\bottomrule
\end{tabular}
\label{table:methods}
\vskip -0.2in
\end{table*}

\subsection{Truth Methods}

\textit{Truth Methods} are methods designed to estimate the truthfulness or correctness of an LLM’s response to a given query. These methods operate in an \textit{off-the-shelf} manner, meaning they do not interfere with the generation process itself, but instead assign a post hoc truthfulness score (referred to as a \textit{truth value}) after the response has been produced. Each Truth Method can optionally define its own parameters and must implement a standardized \texttt{forward} function, which takes as input the relevant generation-time information, such as generated token ids, the LLM and tokenizer objects, and returns a truth value. All methods inherit from the \texttt{TruthMethod} base class and follow a consistent interface, making the library easily extensible for users to implement custom methods.

Truthfulness estimation can be approached in a variety of ways, each with distinct trade-offs. The first major axis of variation is the use of external context: for example, Natural Language Inference (NLI)~\cite{lei2023chainnaturallanguageinference} methods assess truthfulness relative to supporting documents, while Uncertainty Quantification (UQ) methods rely solely on the model’s output probabilities or internal states and do not require any external resources. In addition to document-grounding (1), we categorize truth methods along three further dimensions: (2) whether the method is supervised or self-supervised, that is, whether it requires separate training or can operate in a zero-shot fashion; (3) the level of access to the underlying model, ranging from black-box (output only), to gray-box (output probabilities), to white-box (internal representations); and (4) whether the method requires sampling, some approaches explore the output space to assign truth values, while others operate directly on a single response, which often reflects their computational cost. A representative subset of currently available methods and their categorization is provided in Table~\ref{table:methods}.

\subsection{Unified Generation Interface}

TTLM provides a unified generation interface that supports both locally hosted models via HuggingFace and API-based models through LiteLLM. This interface enables seamless integration of truth prediction with model inference, regardless of deployment type. The core function, \texttt{generate\_with\_truth\_value}, accepts a chat history formatted as a list of message dictionaries (including system prompts, user queries, and prior exchanges), along with a list of predefined truth methods. It also supports standard generation parameters such as temperature, sampling strategy, and maximum token limits, with full compatibility with both HuggingFace and LiteLLM generation arguments.

The function returns the generated output alongside the assigned truth values for each specified truth method and each truth methods' specific details if desired. This streamlined interface enables effortless evaluation of generation reliability across diverse model backends. Figure~\ref{fig:overview}.1 illustrates the function’s architecture, and a code example is shown below.

\begin{lstlisting}[language=Python,
caption={Usage of \texttt{generate\_with\_truth\_value}}]
import TruthTorchLM as ttlm
# Define truth methods
lars = ttlm.truth_methods.LARS()
confidence = ttlm.truth_methods.Confidence()
self_detection = ttlm.truth_methods.SelfDetection(number_of_questions=5)
truth_methods = [lars, confidence, self_detection]

# Define chat input
chat = [{"role": "system", "content": "You are a helpful assistant."},
    {"role": "user", "content": "What is the capital city of France?"}]

# Generate with a HuggingFace model
output_hf_model= ttlm.generate_with_truth_value(
    model=model, tokenizer=tokenizer,
    messages=chat,
    truth_methods=truth_methods,
    max_new_tokens=100, temperature=0.7)

# Generate with an API-based model
output_api_model=ttlm.generate_with_truth_value(
    model="GPT-4o", messages=chat,
    truth_methods=truth_methods)
\end{lstlisting}

\subsection{Evaluation of Truth Methods}

Truth methods assign a scalar score, referred to as the \textit{truth value}, to each model-generated output or individual claim. In short-form question answering tasks, evaluation follows a simple principle: if the generation is correct with respect to the ground truth, the assigned truth value should be high; if incorrect, it should be low. We explain the evaluation in long-form generations in Section \ref{long-form}.

Since free-form generations may vary lexically even when correct, we employ both traditional and modern correctness evaluators. Classical approaches include string-based metrics such as ROUGE~\cite{lin-2004-rouge}, Exact Match, and BLEU~\cite{papineni-etal-2002-bleu}, while more recent methods such as \textit{Model-as-a-Judge} leverage large language models to assess semantic correctness\cite{lin2023generating, yaldiz2024designlearntrainablescoring}. \texttt{TruthTorchLM} supports all of these evaluation criteria out-of-the-box. Once correctness labels are assigned, the performance of a truth method can be measured using both threshold-independent metrics, such as AUROC and PRR, and threshold-dependent metrics like F1 score, accuracy, precision, and recall. In Figure~\ref{fig:overview}.4, we provide the design of evaluation functionality and, below is an example illustrating how to run evaluation using TTLM on the TriviaQA dataset:

\begin{lstlisting}[language=Python, caption={Evaluating truth methods on TriviaQA}]
# Define correctness evaluator
model_judge = ttlm.evaluators.ModelJudge('gpt-4o-mini')

# Use built-in or custom datasets for evaluation
results = ttlm.evaluate_truth_method(
    dataset='trivia_qa',
    model=model, tokenizer=tokenizer,
    truth_methods=truth_methods,
    eval_metrics=['auroc', 'prr', 'accuracy'],
    correctness_evaluator=model_judge,
    size_of_data=1000, max_new_tokens=64)
\end{lstlisting}

\subsection{Calibration of Truth Methods}\label{calibration}

Different truth methods may produce scores on varying ranges. For example, some methods output values between 0 and 1, while others produce unbounded negative scores (e.g., in the range $(-\infty, 0]$). As a result, directly comparing or interpreting these raw truth values can be challenging.

To address this, TTLM supports the calibration of truth method outputs. Calibration maps the original score range into a normalized interval, typically $[0, 1]$, where 0 represents minimal likelihood of truthfulness and 1 represents maximal likelihood. This enables both meaningful comparison across methods and the possibility of ensembling multiple truth scores into a unified signal, as demonstrated in prior work~\cite{bakman2025reconsideringllmuncertaintyestimation}.

We provide several calibration techniques, including Isotonic Regression~\cite{Han2017IsotonicRI}, and simple min-max normalization. Some calibration methods require labeled data for supervision, while others can operate in an unsupervised manner using only queries. Figure~\ref{fig:overview}.3 provides the system overview of the calibration feature.

\subsection{Predicting Truthfullness in Long Form Generation}\label{long-form}

Most questions require long-form generations that contain multiple factual claims, some correct, others incorrect. Evaluating the truthfulness of such outputs is non-trivial. Assigning a single truthfulness score to the entire generation lacks granularity and is not intuitive. To address this, we assign truth values to each individual factual claim within the generation, a strategy also adopted in prior work \cite{semantic-nature, wei2024longformsafe, fadeeva-etal-2024-tokenlevel, zhang-etal-2024-luq, min-etal-2023-factscore}.

To extract individual claims from a generation, the long-form text must first be decomposed. The quality of the decomposition process is critical for reliable truthfulness assessment. Each extracted claim must be self-contained and contextually coherent to enable accurate evaluation. TTLM's \textit{Decomposition Methods} use language models with carefully designed prompts to ensure high-quality results across a wide range of topics.  Users can choose any capable model and optionally enforce a structured output format to prevent parsing issues.

Next step is the assessment of truthfulness of the decomposed claims. However, most truth methods are designed for short-form generations and are not inherently applicable to long-form outputs. To address this limitation, TTLM introduces \textit{Claim Check Methods}. These methods operate on individual claims extracted from long-form generations and assign truth scores to each claim. Similar to truth methods, each claim check method can define its own parameters and must implement a standardized forward function, which takes a claim along with relevant generation-time information and returns a truth value.

Claim check methods serve two main purposes:
1. Wrapper functionality: They adapt existing truth methods for claim-level checks (e.g., claim-specific question generation). In this case, the claim check method is initialized with one or more truth methods as input. TTLM provides three such wrapper methods by default.
2. Claim-level evaluation: These methods are specifically designed for assessing individual claims directly.
All claim check methods inherit from the \texttt{ClaimCheckMethod} base class, ensuring a consistent, extensible interface.

To generate a long-context output with corresponding truth values, TTLM contains \texttt{long\_form\_generation\_with\_truth\_value}. This function accepts a chat history, a set of claim check methods, and a decomposition method. First, it generates a response ant this process is fully compatible with both Hugging Face and LiteLLM, supporting their respective generation configurations. The output is then decomposed into individual factual claims, each of which is evaluated using the specified claim check methods. The function returns the full generation, the set of claims with their associated truth values, and optionally, detailed metadata describing the decomposition and truth assessment processes. Figure~\ref{fig:overview}.2 provides an overview of the long-form generation functionality, with a code example shown below.

\begin{lstlisting}[language=Python, caption={Long-form generation with truth values}]
import TruthTorchLM.long_form_generation as LFG
#define a decomposition method 
decomposition_method = LFG.decomposition_methods.StructuredDecompositionAPI(model="gpt-4o-mini", decomposition_depth=1) 

#claim check method that apply truth methods
qa_generation = LFG.claim_check_methods.QuestionAnswerGeneration(model="gpt-4o-mini", truth_methods=[confidence, lars]) 

#claim check methods designed for this purpose
ac_entailment = LFG.claim_check_methods.AnswerClaimEntailment( model="gpt-4o-mini", num_questions=3, num_answers_per_question=2)

#define a chat history
chat = [{"role": "system", "content": "You are a helpful assistant."},
        {"role": "user", "content": "Who is Ryan Reynolds?"}]

# Generate with an API-based model
out = LFG.long_form_generation_with_truth_value(
        model="gpt-4o-mini", messages=chat, 
        decomp_method=decomposition_method, 
        claim_check_methods=[qa_generation, ac_entailment])
\end{lstlisting}

TTLM evaluates claim check methods, either individually or in combination with truth methods, at the claim level within long-form generations. Since ground truth labels are typically unavailable for claims extracted from long-form outputs, we adopt the SAFE algorithm \cite{wei2024longformsafe}, which estimates claim correctness via Google Search. Once correctness labels are established, each (claim, truth value) pair is treated as a distinct evaluation sample, and assessment is conducted across all claims in the dataset. As in short-form evaluation, both threshold-dependent and threshold-independent metrics can be used to measure performance. A code sample for evaluating long-form generation is included in Appendix~\ref{codes}.


\begin{table*}[ht]
\centering
\caption{AUROC and PRR performance of truth methods on TriviaQA, GSM8K, and FactScore-Bio, across two models: LLaMA-3 8B and GPT-4o-mini.}
\vskip -0.1in
\resizebox{\textwidth}{!}{%
\begin{tabular}{l|cc|cc|cc||cc|cc|cc}
\toprule
 & \multicolumn{6}{c||}{\textbf{LLaMA-3 8B}} & \multicolumn{6}{c}{\textbf{GPT-4o-mini}} \\
\cmidrule{2-13}
 & \multicolumn{2}{c|}{TriviaQA} & \multicolumn{2}{c|}{GSM8K} & \multicolumn{2}{c||}{FactScore-Bio}
 & \multicolumn{2}{c|}{TriviaQA} & \multicolumn{2}{c|}{GSM8K} & \multicolumn{2}{c}{FactScore-Bio} \\
 \textbf{Truth Methods}& \textbf{AUROC} & \textbf{PRR} & \textbf{AUROC} & \textbf{PRR} & \textbf{AUROC} & \textbf{PRR}
 & \textbf{AUROC} & \textbf{PRR} & \textbf{AUROC} & \textbf{PRR} & \textbf{AUROC} & \textbf{PRR} \\
\midrule
LARS                    & 0.861 & 0.783 & 0.834 & 0.719 & 0.677 & 0.391 & 0.852 & 0.766 & 0.840 & 0.686 & 0.640 & 0.294 \\
MARS                    & 0.763 & 0.635 & 0.730 & 0.488 & 0.660 & 0.367 & 0.792 & 0.668 & 0.735 & 0.480 & 0.655 & 0.405 \\
SelfDetection           & 0.780 & 0.590 & 0.556 & 0.090 & 0.687 & 0.369 & 0.799 & 0.587 & 0.736 & 0.421 & 0.671 & 0.313 \\
PTrue                   & 0.727 & 0.485 & 0.654 & 0.307 & 0.670 & 0.368 & 0.772 & 0.509 & 0.833 & 0.636 & 0.658 & 0.372 \\
AttentionScore          & 0.523 & 0.092 & 0.503 & -0.024 & 0.644 & 0.263 & -- & -- & -- & -- & -- & -- \\
CrossExamination        & 0.664 & 0.377 & 0.585 & 0.187 & 0.683 & 0.361 & 0.718 & 0.483 & 0.768 & 0.551 & 0.635 & 0.289 \\
Eccentricity            & 0.809 & 0.645 & 0.703 & 0.450 & 0.695 & 0.415 & 0.817 & 0.632 & 0.754 & 0.455 & 0.671 & 0.421 \\
GoogleSearchCheck       & 0.672 & 0.470 & -- & -- & -- & -- & 0.779 & 0.673 & -- & -- & -- & -- \\
Inside                  & 0.711 & 0.478 & 0.689 & 0.354 & 0.636 & 0.221 & -- & -- & -- & -- & -- & -- \\
KernelLanguageEntropy   & 0.792 & 0.596 & 0.662 & 0.296 & 0.680 & 0.396 & 0.820 & 0.635 & 0.706 & 0.349 & 0.678 & 0.397 \\
SAPLMA                  & 0.850 & 0.726 & 0.815 & 0.642 & 0.651 & 0.347 & -- & -- & -- & -- & -- & -- \\
SemanticEntropy         & 0.799 & 0.652 & 0.699 & 0.417 & 0.682 & 0.403 & 0.813 & 0.673 & 0.735 & 0.464 & 0.681 & 0.447 \\
MultiLLMCollab          & 0.632 & 0.350 & 0.689 & 0.320 & 0.681 & 0.347 & 0.778 & 0.565 & 0.933 & 0.879 & 0.671 & 0.399 \\
SAR                     & 0.804 & 0.679 & 0.768 & 0.590 & 0.674 & 0.389 & 0.835 & 0.724 & 0.764 & 0.512 & 0.671 & 0.433 \\
VerbalizedConfidence    & 0.759 & 0.547 & 0.579 & 0.234 & 0.698 & 0.460 & 0.836 & 0.740 & 0.652 & 0.369 & 0.717 & 0.514 \\
DirectionalEntailmentGraph 
                        & 0.745 & 0.513 & 0.731 & 0.501 & 0.659 & 0.347 & 0.778 & 0.532 & 0.736 & 0.439 & 0.658 & 0.380 \\
\bottomrule
\end{tabular}
}
\label{table:results}
\vskip -0.1in
\end{table*}

\section{Related Works}

The most closely related open-source libraries to \texttt{TruthTorchLM} are \texttt{GuardrailsAI}~\cite{guardrails2024} and \texttt{LM-Polygraph}~\cite{fadeeva-etal-2023-lm}. \texttt{GuardrailsAI} implements guardrail mechanisms for safe and structured LLM outputs, primarily through document-grounded verification. \texttt{LM-Polygraph}, on the other hand, focuses on uncertainty quantification methods for generative language models.\texttt{TruthTorchLM} distinguishes itself from both in an important way. TTLM is explicitly designed for truthfulness prediction and aims to unify a wide spectrum of methods, ranging from uncertainty-based to supervised, document-grounded, and LLM-collaboration approaches. In contrast, \texttt{GuardrailsAI} is limited to document-grounded verification, while \texttt{LM-Polygraph} covers only uncertainty-based techniques, which represent a subset of the methods included in TTLM.

\section{Experiments}

We evaluate the performance of a subset of available truth methods listed in Table~\ref{table:methods} using our proposed library, \texttt{TruthTorchLM}. In this section, we present the details of our experimental setup and provide a discussion of the results.

\paragraph{Datasets}
Our primary evaluation focuses on short-form question answering, a standard benchmark for assessing truthfulness. We use 1,000 samples from TriviaQA~\cite{joshi2017triviaqa} and GSM8K~\cite{cobbe2021gsm8k} for open-ended and mathematical reasoning questions, respectively. For long-form evaluation, we use FactScore-Bio~\cite{min-etal-2023-factscore}, which targets biographical questions with multi-fact generations.

\paragraph{Models}
We conduct evaluations using both open- and closed-weight language models. Specifically, we use \texttt{LLaMA-3-8B}~\cite{llama3modelcard}, an open-source model that enables full access to internal states, and \texttt{GPT-4o-mini}~\cite{openai2023gpt4}, a closed-weight API model. Note that white-box truth methods are not applicable to \texttt{GPT-4o-mini}.

\paragraph{Metrics}
As discussed in Section~\ref{calibration}, different truth methods produce scores on varying numerical scales, which complicates the use of fixed thresholds for evaluation. While threshold-based metrics such as accuracy can be informative, they require method-specific thresholds, introducing potential bias or instability in comparison.

To mitigate this issue, we primarily report threshold-free metrics, following prior work~\cite{kuhn2023semantic, bakman2025reconsideringllmuncertaintyestimation}. Specifically, we use the Area Under the Receiver Operating Characteristic Curve (AUROC) and the Prediction Rejection Ratio (PRR). AUROC measures a method’s ability to distinguish between truthful and untruthful outputs across all possible thresholds, with values ranging from 0.5 (random performance) to 1.0 (perfect discrimination). PRR quantifies the relative precision gain obtained by rejecting low-confidence predictions and ranges from 0.0 (random rejection) to 1.0 (perfect rejection).


\paragraph{Correctness Measure}
Since our tasks involve free-form generation, the model outputs may be semantically correct even if they do not lexically match the ground truths. To account for this, we adopt the \textit{LLM-as-a-judge} paradigm, following prior work~\cite{bakman2025reconsideringllmuncertaintyestimation, semantic-nature}. Specifically, we prompt a language model with the question, the generated answer, and the reference answer, and ask it to provide a binary correctness judgment (0 or 1). For long-form generations in FactScore-Bio, where reference ground truths are unavailable  we use the SAFE algorithm~\cite{wei2024longformsafe} to automatically extract and assess the correctness of individual factual claims within the generated text.

\subsection{Discussion}

The results are summarized in Table~\ref{table:results}. Since each method entails different trade-offs, such as computational overhead, model access level, and supervision requirements, their performance varies accordingly. In short-form QA tasks (TriviaQA and GSM8K), LARS and SAPLMA achieve the highest performance, except on GSM8K with GPT-4o-mini, which is expected given that both are trained on labeled data. Among self-supervised methods, SAR  performs best on both TriviaQA and GSM8K for the LLaMA-3-8B model. For GPT-4o-mini, Verbalized Confidence achieves the best results on TriviaQA, while MultiLLMCollab leads on GSM8K. 

FactScore-Bio evaluates long-form generation, which typically involves multiple factual claims and thus presents a more challenging setting for truthfulness detection. On this task, performance generally drops across methods compared to short-form QA. Verbalized confidence achieves the best results on both models. Eccentiricity and Semantic Entropy performs next best as sampling based methods, with Semantic Entropy showing stronger results for GPT-4o-mini. 

\section{Conclusion}
In this work, we introduced \texttt{TruthTorchLM}, an open-source library for evaluating and developing truthfulness prediction methods for large language models. TTLM unifies a diverse set of techniques under a common interface, supports both short- and long-form generation tasks, and includes tools for evaluation, calibration, and extensibility. We hope TTLM serves as a valuable resource for the community and accelerates research in building more trustworthy and reliable language models.

\section*{Ethics Statement}

We acknowledge the ethical considerations associated with the development and release of truthfulness prediction tools for large language models (LLMs). Our library, TruthTorchLM, is designed to assist researchers and practitioners in systematically evaluating and improving the truthfulness of LLM outputs. It does not generate content on its own; any harmful or incorrect content produced by language models is not the product of this library. Our goal is to help detect and reduce untruthful outputs.

All experiments in this work were conducted on publicly available datasets (TriviaQA, GSM8K, and FactScore-Bio) that do not contain personally identifiable or sensitive information. No private or user-generated data was collected or used during development or evaluation. We encourage responsible use of our library and caution that automated truthfulness prediction should complement, not replace, human oversight, especially in high-stakes domains such as health, law, and finance.


\bibliography{main}

\clearpage
\appendix
\section{Datasets Statistics}

TriviaQA contains question-answer pairs authored by trivia enthusiasts. Among the ~17.2k samples in the test split, we use a random subset of 1000 samples in our evaluations. GSM8K is composed of grade school math word problems. It contains 1.32k samples in the test set. Similar to GSM8K, we use a subset of 1000 samples in our experiments. Lastly, FactScore-Bio contains biography queries about people from Wikipedia. We used a random subset of 50 questions from this dataset. After generation decomposition, the total number of claims is 1290 for GPT-4o-mini and 1764 for Llama-3-8B. We provide sample questions from each dataset in Table~\ref{tab:data_samples}.

\begin{table*}[!htbp]
\vskip 0.3in

\centering
\fontsize{4.2}{3.4}\selectfont
\resizebox{\textwidth}{!}{%
\begin{tabular}{c|p{5cm}|c}
\toprule
&\multicolumn{1}{c|}{ \textbf{Question}} &\multicolumn{1}{c}{ \textbf{Ground Truth}}  \\
\midrule[\heavyrulewidth]

\multirow{8}{*}{\rotatebox{90}{\textbf{TriviaQA}}}
&\multicolumn{1}{m{4cm}|}{David Lloyd George was British Prime Minister during the reign of which monarch?}&
King George V\\
\cmidrule{2-3}
&\multicolumn{1}{m{4cm}|}{How many symphonies did Jean Sibelius compose?}&
Seven\\
\cmidrule{2-3}
&\multicolumn{1}{m{4cm}|}{The capital of Brazil was moved from Rio de Janeiro to the purpose-built capital city of Brasilia in what year?}&
1960\\

\midrule[\heavyrulewidth]
\multirow{13}{*}{\rotatebox{90}{\textbf{GSM8K}}}
&\multicolumn{1}{m{4cm}|}{Natalia sold clips to 48 of her friends in April, and then she sold half as many clips in May. How many clips did Natalia sell altogether in April and May?}&72\\
\cmidrule{2-3}
&\multicolumn{1}{m{4cm}|}{Julie is reading a 120-page book. Yesterday, she was able to read 12 pages and today, she read twice as many pages as yesterday. If she wants to read half of the remaining pages tomorrow, how many pages should she read?}&
42\\
\cmidrule{2-3}
&\multicolumn{1}{m{4cm}|}{Mr. Sam shared a certain amount of money between his two sons, Ken and Tony. If Ken got \$1750, and Tony got twice as much as Ken, how much was the money shared?}&
5250\\

\midrule[\heavyrulewidth]
\multirow{6}{*}{\rotatebox{90}{\textbf{FactScore-Bio}}}
&\multicolumn{1}{m{4cm}|}{Tell me a bio of Vaira Vīķe-Freiberga.}&-\\
\cmidrule{2-3}
&\multicolumn{1}{m{4cm}|}{Tell me a bio of Ji Sung.}&
-\\
\cmidrule{2-3}
&\multicolumn{1}{m{4cm}|}{Tell me a bio of Baltasar Corrada del Río.}&
-\\
\cmidrule{2-3}
&\multicolumn{1}{m{4cm}|}{Tell me a bio of Henry Santos.}&
-\\

\bottomrule
\end{tabular}}
\caption{Data samples from the datasets we use in our evaluations: TriviaQA, GSM8K, and FactScore-Bio }
\label{tab:data_samples}
\end{table*}

\section{Additional Code Snippets}\label{codes}

Below is an example illustrating how to calibrate a set of truth methods on the TriviaQA dataset:

\begin{lstlisting}[language=Python, caption={Calibrating multiple truth methods using Isotonic Regression.}]
# Assign a calibrator to each method
for truth_method in truth_methods:
    truth_method.set_normalizer(ttlm.normalizers.IsotonicRegression())

# Calibrate using labeled evaluation data
calib_results = ttlm.calibrate_truth_method(
    dataset='trivia_qa',
    model=model, tokenizer=tokenizer,
    truth_methods=truth_methods,
    correctness_evaluator=model_judge,
    size_of_data=1000, max_new_tokens=64)
\end{lstlisting}

We provide a code sample below that evaluates truth methods in long-form generation setting:

\begin{lstlisting}[language=Python, caption={Evaluation on long-form generation.}]
#Define claim evaluator 
safe = LFG.ClaimEvaluator(
    rater='gpt-4o-mini', 
    tokenizer = None, 
    max_steps = 5, 
    max_retries = 10, 
    num_searches = 3) 

# Use built-in or custom datasets for evaluation
results = LFG.evaluate_truth_method_long_form(
    dataset='longfact_objects', 
    model=model, tokenizer=tokenizer,
    sample_level_eval_metrics=['f1'],
    dataset_level_eval_metrics=['auroc', 'prr'],
    decomp_method=decomposition_method, 
    claim_check_methods=[qa_generation],
    claim_evaluator = safe)

\end{lstlisting}

\section{Generation Decomposition}

We provide a sample output of our decomposition process in Table~\ref{tab:decomp_sample2}. The model used for decomposition is GPT-4o-mini and the output structure is forced by instructor\footnote{\url{https://python.useinstructor.com/}}.

\begin{table*}[ht]
\centering
\begin{tabular}{|p{\textwidth}|}
\hline
\textbf{Question:} Tell me a bio of Andrew Shue.

\textbf{Generation:}  Andrew Shue is an American actor, producer, and former professional soccer player. He was born on January 20, 1967, in South Orange, New Jersey, USA.

Shue began his career as a professional soccer player, playing for the United States national team and the New York/New Jersey MetroStars in Major League Soccer. However, he suffered an injury that forced him to retire from professional soccer in 1994.

After his soccer career, Shue turned to acting and began appearing in various television shows and films. Some of his notable roles include:

* "The Breakfast Club" (1985) - Shue played the role of Andrew Clark, a popular high school student.

* "Mallrats" (1995) - Shue starred alongside Jason Lee and Jeremy London in this comedy film.

* "The Adventures of Pete \& Pete" (1992-1996) - Shue played the role of C.J. Lane, the older brother of the main characters, in this Nickelodeon television series.

* "Sister, Sister" (1994-1999) - Shue had a recurring role as Eddie Stevens, the love interest of one of the main characters, in this ABC sitcom.

Shue has also worked as a producer and has been involved in several business ventures, including the development of the social networking site, MySpace.

Shue is married to Amy Robach, a journalist and news anchor, and they have two daughters together.

\textbf{Claims:}  

[`Andrew Shue is an American actor.',\\
  `Andrew Shue is a producer.',\\
  `Andrew Shue is a former professional soccer player.',\\
  `Andrew Shue was born on January 20, 1967.',\\
  `Andrew Shue was born in South Orange, New Jersey, USA.',\\
  `Andrew Shue began his career as a professional soccer player.',\\
  `Andrew Shue played for the United States national soccer team.',\\
  `Andrew Shue played for the New York/New Jersey MetroStars in Major League Soccer.',\\
  `Andrew Shue suffered an injury that forced him to retire from professional soccer in 1994.',\\
  `Andrew Shue turned to acting after his soccer career.',\\
  `Andrew Shue began appearing in various television shows.',\\
  `Andrew Shue began appearing in films.',\\
  `Andrew Shue played the role of Andrew Clark in "The Breakfast Club" in 1985.',\\
  `Andrew Shue starred in "Mallrats" in 1995.',\\
  `Andrew Shue starred alongside Jason Lee in "Mallrats".',\\
  `Andrew Shue starred alongside Jeremy London in "Mallrats".',\\
  `From 1992 to 1996, Andrew Shue played the role of C.J. Lane in "The Adventures of Pete \& Pete".',\\
  `From 1994 to 1999, Andrew Shue had a recurring role as Eddie Stevens in "Sister, Sister".',\\
  `Andrew Shue has worked as a producer.',\\
  `Andrew Shue has been involved in several business ventures.',\\
  `Andrew Shue has been involved in the development of the social networking site MySpace.',\\
  `Andrew Shue is married to Amy Robach.',\\
  `Amy Robach is a journalist.',\\
  `Amy Robach is a news anchor.',\\
  `Andrew Shue and Amy Robach have two daughters together.'] \\

\hline
\end{tabular}
\caption{Output of long-text decomposition. The question is from FactScore-Bio and the model used to generate the answer is LLaMa-3-8B.}
\label{tab:decomp_sample2}
\end{table*}
\end{document}